\pdfoutput=1

\documentclass[11pt]{article}

\usepackage[preprint]{acl}

\usepackage{times}
\usepackage{latexsym}

\usepackage[T1]{fontenc}

\usepackage[utf8]{inputenc}

\usepackage{microtype}

\usepackage{inconsolata}

\usepackage{graphicx}
\usepackage{algorithm}
\usepackage{algpseudocode}
\usepackage{amsmath}
\usepackage{booktabs}
\usepackage{multirow}
\usepackage{float}
\usepackage{stfloats}
\usepackage{array}    
\title{BloomWise: Enhancing Problem-Solving capabilities of Large Language Models using Bloom's-Taxonomy-Inspired Prompts}

\author{
Maria-Eleni Zoumpoulidi\textsuperscript{2} \quad 
Georgios Paraskevopoulos\textsuperscript{2} \quad 
Alexandros Potamianos\textsuperscript{1} \\
\texttt{m.zoumpoulidi@athenarc.gr} \quad
\texttt{g.paraskevopoulos@athenarc.gr} \quad
\texttt{potam@central.ntua.gr} \\
\textsuperscript{1}Speech and Language Processing Group, National Technical University of Athens, Greece \\
\textsuperscript{2}Institute for Language and Speech Processing, Athena Research Center, Greece
}

\begin{document}
\maketitle
\begin{abstract}
Despite the remarkable capabilities of large language models (LLMs) across a range of tasks, mathematical reasoning remains a challenging frontier. Motivated by the observation that humans learn more effectively when prompted not what to think but how to think, we introduce BloomWise, a cognitively-inspired prompting technique designed to enhance LLMs' performance on mathematical problem solving while making their solutions more explainable. BloomWise encourages LLMs to generate solutions - in the form of explanations - by progressing through a sequence of cognitive operations—from basic (e.g., remembering) to more advanced reasoning skills (e.g., evaluating)—mirroring how humans build understanding. The process iterates through these levels, halting early if a convergence criterion is met: specifically, if two or more consecutive levels yield the same answer, the solution from the earliest such level is output; otherwise, the process continues until all levels are completed. Through extensive experiments across five popular math reasoning datasets, we demonstrate the effectiveness of BloomWise. We also present comprehensive ablation studies to analyze the strengths of each component within our system. 
\end{abstract}

\section{Introduction}
Mathematical reasoning has long been regarded as a pinnacle of human intellect—demanding abstraction, logical structure, and creativity. As LLMs achieve remarkable fluency in natural language, empowering them with robust mathematical reasoning skills is crucial for scientific and technological progress. Yet, mastering the complexity and nuance of mathematical problem solving remains a formidable challenge for LLMs, motivating new approaches.

Numerous research efforts have leveraged in-context learning (ICL) \citep{DBLP:conf/nips/BrownMRSKDNSSAA20} to improve the problem-solving capabilities of LLMs. Some of the most widely used techniques involve encouraging the LLM through prompts to develop a textual rationale with Chain-of-Thought prompting \citep{DBLP:conf/nips/Wei0SBIXCLZ22} (CoT) or Python functions with Program-Aided Language Model \citep{DBLP:journals/corr/abs-2211-10435} and Program-of-Thought prompting \citep{DBLP:journals/corr/abs-2211-12588} (PAL or PoT). Each of these methods comes with its own strengths and limitations: CoT enables flexible, sequential narrative-style reasoning but often struggles with precise numerical computation \citep{DBLP:conf/nips/Wei0SBIXCLZ22}, \citep{lewkowycz2022solving}, while code-based approaches like PoT and PAL offer accurate calculations via Python interpreters but lack the ability to handle unknown variables.

For this reason, research efforts have pivoted towards integrating multiple methodologies, aiming to identify the most suitable approach for each specific problem, while harnessing the collective strengths of various techniques. One such method, X of Thoughts \citep{liu-etal-2023-plan}, selects from CoT, PoT, or EoT (Equations of Thought) depending on the nature of the problem, applies the selected approach, and verifies the result (iteratively, until a correct solution is reached).

Building on the integration of diverse reasoning strategies—and aiming for more structured, human-aligned, and explainable mathematical problem solving in LLMs—we introduce BloomWise, a novel cognitively inspired prompting method. BloomWise guides LLMs to generate solutions in the form of explanations by methodically engaging higher-order cognitive functions in a hierarchical manner, persisting through the process until the correct solution is reached via a convergence criterion: if two or more consecutive levels yield the same result, the process halts and outputs the solution from the earliest such level. The motivation behind this approach is that, while encouraging LLMs to follow a specific methodology or approach can be effective, prompting them how—rather than what—to think enables more in-depth processing. 

To demonstrate the effectiveness of our approach, we conduct extensive experiments on five popular mathematical reasoning datasets and achieve consistent improvements. Additionally, we explore several variants of our method, including majority voting among
levels, and Program of Bloom, that combines Bloom prompting with PoT. 

The main contributions  are: 
\begin{enumerate}
    \item 
We introduce a novel cognitively-inspired multi-level prompting method for solving mathematical (and potentially other types of) problems based on Bloom's taxonomy, combining robust reasoning and enhanced explainability. Our code will be available to the research community under the Apache 2.0 license\footnote{\tiny{\url{https://osf.io/rnx7j/?view_only=0758fad9d5164bbb865f9ca91a2a480d}}}.
\item We incorporate the idea of early stopping in prompting, motivated by recent work on dynamically adjusting test-time compute during inference (\citet{snell2025scaling}, \citet{manvi2024adaptiveinferencetimecomputellms}): the execution of our method terminates before iterating through all levels of Bloom's taxonomy when a correct solution is reached.
\item We investigate the performance of various LLMs at each cognitive level of the Bloom taxonomy for five popular math  datasets. 
\end{enumerate}
Our results offer valuable insights into the cognitive skills exhibited by each LLM, as well as the skills required to solve different types of mathematical problems. Furthermore, we demonstrate the potential of muti-level cognitively-inspired prompting for improving accuracy and enhancing explainability. 

\section{Related Work}
As Large Language Models (LLMs) continue to advance, an array of prompting strategies has emerged to strengthen their reasoning abilities. Early progress was made through chain-of-thought prompting, which encourages step-by-step reasoning \citep{DBLP:conf/nips/Wei0SBIXCLZ22}, utilizing programming to address procedural challenges \citep{DBLP:journals/corr/abs-2211-10435, DBLP:journals/corr/abs-2211-12588} and employing zero-shot prompts that rely on a single guiding sentence to elicit complex responses \citep{DBLP:conf/nips/KojimaGRMI22}.  Furthermore, the Tree-of-Thoughts approach \citep{DBLP:journals/corr/abs-2305-10601} navigates through various reasoning pathways and traverses tree-like structures of reasoning states. Moreover, X of Thoughts \citep{liu-etal-2023-plan} selects, applies and verifies the most suitable among the techniques of CoT (Chain of Thought), PoT (Program of Thought), and EoT (Equations of Thought) iteratively, until a correct solution is reached.

\section{Preliminaries: Bloom's Taxonomy}
\begin{figure}[htb!]
    \centering
    \includegraphics[width=0.7\linewidth]{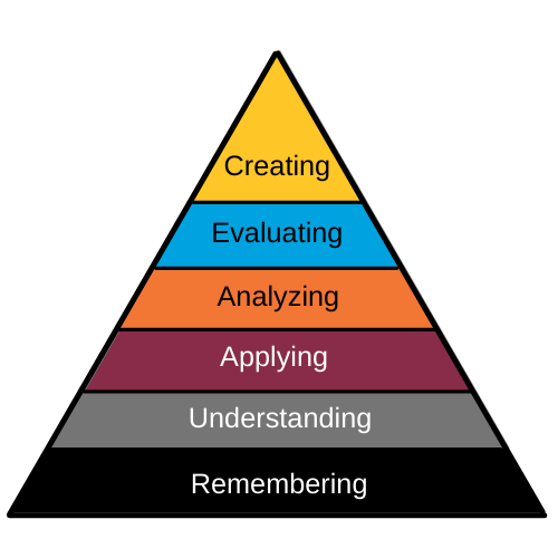}
    \caption{Bloom's Taxonomy (as revised  by \citet{anderson2001taxonomy}).}
    \label{fig:blooms_taxonomy}
    \vspace{-1em}
\end{figure}

Bloom’s Taxonomy provides a hierarchical classification of thinking according to six levels of cognitive complexity. The original model, introduced in the 1950s, organizes cognitive processes based on the following order: remembering, understanding, applying, analyzing, synthesizing, and evaluating. The taxonomy is hierarchical, as shown in Fig.~\ref{fig:blooms_taxonomy}, where each level is subsumed by the higher levels. In 2001, the taxonomy was revised by \citet{anderson2001taxonomy}, resulting in a new sequence: remembering, understanding, applying, analyzing, evaluating, and creating. Our work is based on the revised taxonomy of \citet{anderson2001taxonomy}.
The steps used in the Taxonomy are defined as follows \citep{forehand2005blooms}:\\
\noindent{\bf Remembering:} 
Retrieving, recognizing, and recalling relevant knowledge from long-term memory.\\
\noindent{\bf Understanding:} 
Constructing meaning from oral, written, and graphic messages through interpreting, exemplifying, classifying, summarizing, inferring, comparing, and explaining.\\
\noindent{\bf Applying:} 
Carrying out or using a procedure through executing, or implementing.\\
\noindent{\bf Analyzing:}
Breaking material into constituent parts, determining how the parts relate to one another and to an overall structure or purpose through differentiating, organizing, and attributing.\\
\noindent{\bf Evaluating:}
Making judgments based on criteria and standards through checking and critiquing.\\
\noindent{\bf Creating:}
Putting elements together to form a coherent or functional whole; reorganizing elements into a new pattern or structure through generating, planning, or producing.
\begin{figure*}[htbp]
    \centering
    \includegraphics[width=\textwidth]{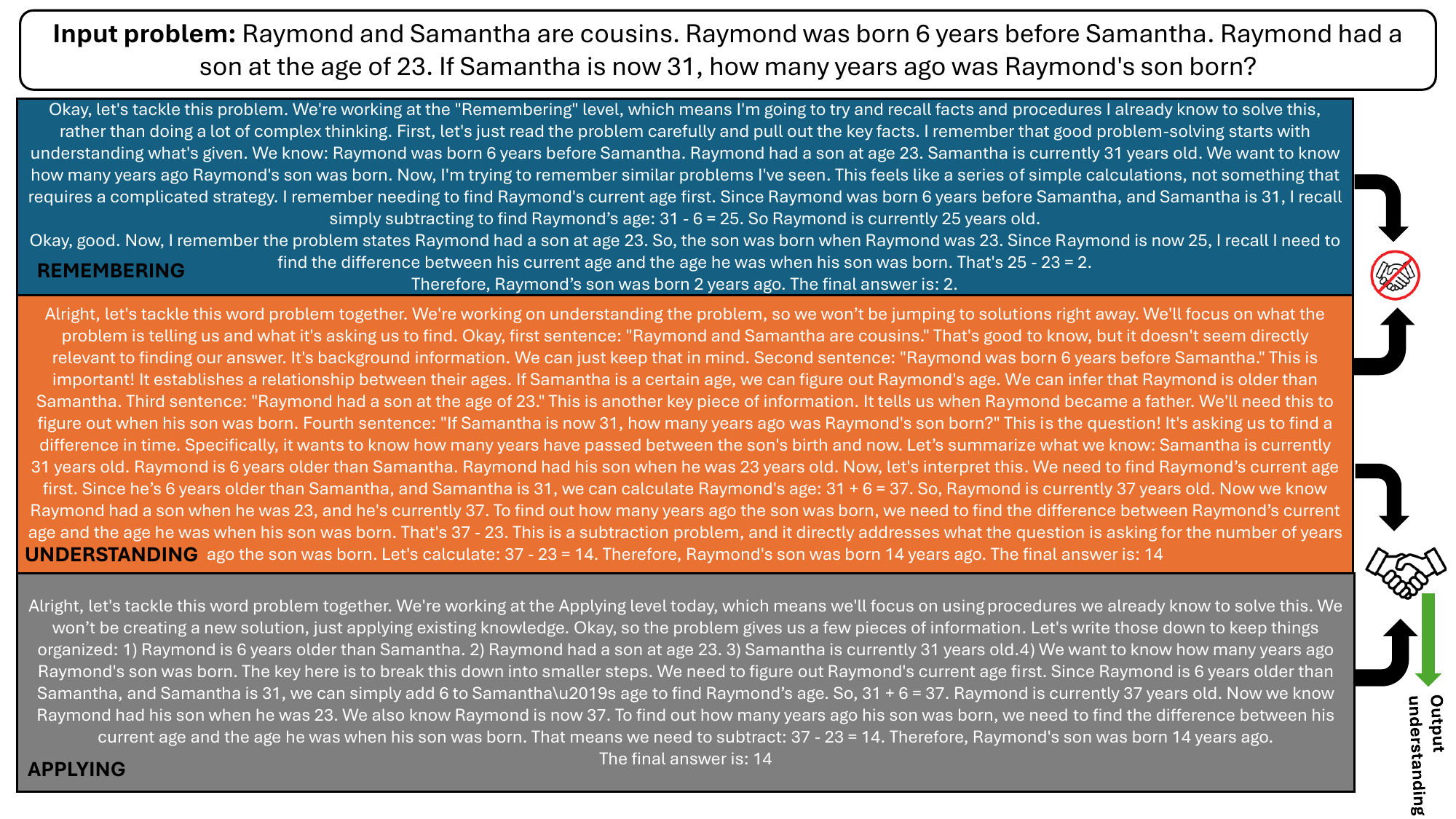}
    \caption{Overview of BloomWise Early Stop algorithm. The math problem is shown on top, followed by the output at each Bloom taxonomy level (blue box, orange and gray box). Input prompts are not shown, please refer to Appendix~\ref{sec:appendix}. The model generates responses corresponding to the first 2 levels of Bloom's taxonomy (blue and orange boxes, Remembering and Understanding respectively). Since there is no consensus between the two, it generates an answer corresponding to the next level (Applying, gray box). Now, there is consensus between the 2 consecutive levels (Understanding and Applying), so the process halts and the response corresponding to the earliest such level (Understanding) is returned.}
    \label{fig:example_bloomwise_overview}
\end{figure*}
\section{BloomWise Early Stop (BLES)}
Our goal is to develop a generalized problem solving framework that iterates through prompts, each corresponding to a level of Bloom’s taxonomy, until the correct solution is reached. Next, we describe the overall framework and introduce each module in detail.

\subsection{The Framework}
In this framework, the system progresses through the levels of Bloom's taxonomy in sequence for a given problem. At each level, the LLM is provided with the problem along with a prompt specifically designed for that level. The model then generates a response—an explanatory solution. In case of consensus between two consecutive levels' numerical results, the process concludes successfully at the earliest of the two levels, and no further levels are explored.  Otherwise, the process continues to the next level until either consensus is reached or all levels have been exhausted. The overall pipeline is described in Algorithm ~\ref{alg:bloomwise} and an overview of the framework can be found in Figure ~\ref{fig:example_bloomwise_overview}.
\begin{algorithm}
\caption{BloomWise Early Stop (BLES)}
\label{alg:bloomwise}
\begin{algorithmic}[1]
\State \textbf{Input:} $i$
\State $results \gets [\,]$
\For{$level$ in $levels$}
    \State $res \gets \text{Reasoning}(level\_prompt, i)$
    \State \text{Append } $res$ \text{ to } $results$
    \If{$|results| \geq 2$ \textbf{and} $results[-1] = results[-2]$}
        \State \textbf{break}
    \EndIf
\EndFor
\If{there exists $k$ such that $results[k] = results[k+1]$}
    \State \Return $results[\min\{k : results[k] = results[k+1]\}]$
\Else
    \State \Return $results[-1]$ \Comment{If no convergence, return the last result}
\EndIf
\end{algorithmic}
\end{algorithm}
\subsection{Prompts}

We designed prompts corresponding to each level of Bloom's Taxonomy: 

\vspace*{2mm}
\noindent 
{\bf Remembering:} 
The model is prompted to solve the problem by retrieving, recognizing, and recalling relevant math facts, formulas, definitions, similar problems or the exact same problem from memory.

 \vspace*{2mm}
 \noindent 
{\bf Understanding:}
The model is prompted to solve the problem by constructing meaning from the problem statement and relevant math concepts and demonstrate the thinking process by interpreting, exemplifying, classifying, summarizing, inferring, comparing, and explaining the concepts involved and what the problem is asking for.

\vspace*{2mm}
 \noindent 
{\bf Applying:} 
The model is prompted to solve the problem by carrying out or using a known procedure.

\vspace*{2mm}
 \noindent 
{\bf Analyzing:}
The model is prompted to solve the problem by breaking it into parts, determining how the parts relate to one another, and identifying patterns or relationships. 

\vspace*{2mm}
 \noindent 
{\bf Evaluating:}
The model is prompted to solve the problem by making judgments about different approaches or potential solutions. 

\vspace*{2mm}
 \noindent 
{\bf Creating:} 
The model is prompted to solve the problem by putting together elements to form a new solution strategy or structure. 
\\
The detailed prompts are shown in Tables ~\ref{tab:system_prompt} and ~\ref{tab:blooms_prompts} in Appendix ~\ref{sec:appendix}. Examples can be found in Appendix ~\ref{sec:appendixB}. 
\subsection{Convergence}
The verification module determines the correctness of a solution based on a convergence criterion. For each problem, the framework progresses through the levels of Bloom’s taxonomy in sequence. At each level, the LLM is given the problem along with a prompt tailored for that specific cognitive level, and generates an explanatory solution. After each level, the numerical result is compared with this from the previous level. If two consecutive levels yield the same result, this consensus is taken as sufficient evidence of correctness, and the process halts at the earliest such level.  

\subsection{Iteration and Early Stopping}
For each problem, the LLM is guided by prompts crafted to align with the respective levels of Bloom's taxonomy. The model continues this process until it produces a correct solution or all levels have been explored. This approach operates on the premise that prompts associated with higher taxonomy levels encourage the LLM to engage with the problem more deeply than those at lower levels. Moreover, if a correct solution is obtained at a lower level, it suggests that more advanced cognitive effort is unnecessary for that particular problem, making early termination of the process a logical choice. This procedure is illustrated in Figure ~\ref{fig:example_bloomwise_overview}.

\subsection{BloomWise Majority Voting (BLM)}
As an alternative to early stopping, we also explored an  approach where the final output is determined by a majority vote of the outputs corresponding to all levels of Bloom's taxonomy. This strategy utilizes the collective reasoning of multiple cognitive stages. For a given question \( q \), each reasoning stage \( s \in \mathcal{S} \) of Bloom's taxonomy produces a numerical result \( R_s(q) \). The majority vote approach selects the final answer as the value that occurs most frequently among these outputs. $\hat {R}_{\text{majority}}(q) \in {0, 1}$ compared to the gold label.
We define the accuracy under the majority vote setting for \( N \) questions  as:
\begin{equation}
Acc_{majority} = \frac{1}{N} \sum_{q} \hat R(q)  ,
\end{equation}
where \[
\hat{R}(q) = \hat{R}_{\text{majority}}(q).
\]
The majority vote setting represents a consensus-based approach where the model aims to solve a given problem based on the most frequent results from the methods employed. In cases where there is a tie (i.e., two or more answers have the same highest frequency), the first encountered answer among those with the highest frequency is chosen.
\begin{table*}[!t]
\centering
\small
\resizebox{\linewidth}{!}{
\begin{tabular}{llccccc}
\toprule
 & Dataset & CoT & PoT & XoT & BloomWise EarlyStop (BLES) & BloomWise Majority Voting (BLM)\\ 
 \midrule

\multirow{5}{*}{GPT-4o-mini} 
                         & GSM8K     &  93.9 & 88.2  &  93.6 & \underline{94.2} & \textbf{94.9}\\
                         & SVAMP     &  93.4 & 92.0  & 93.3  & \textbf{95.1}  & \underline{95.0} \\
                         & Algebra   & 95.0  & 83.3 & 95.9  & \underline{95.9} & \textbf{96.4} \\ 
                         & GSM-hard  & 52.2  & \textbf{71.6}  & 54.7  & \underline{56.0} & 55.9 \\
                         & aime24    &  \underline{10.0} & 3.3  & \underline{10.0}  & \textbf{13.3} & \textbf{13.3}\\ \midrule

\multirow{5}{*}{LLaMA 3.1 70B} 
                         & GSM8K     &  94.5 &  90.0 & 93.6  & \underline{95.3} & \textbf{95.7} \\
                         & SVAMP     & 92.8  &  94.2 & 94.8  & \textbf{95.0} & \underline{94.9} \\
                         & Algebra   & 95.9  & 77.9  & \underline{96.4}  & \underline{96.4} & \textbf{98.2}\\ 
                         & GSM-hard  &  46.0 & \underline{66.5}  & \textbf{69.9}  & 47.5 & 48.8 \\
                         & aime24    & \textbf{23.3}  & 6.7  & 16.7  & \underline{20.0} & \textbf{23.3}\\ \midrule

\multirow{5}{*}{LLaMA 3.1 8B} 
                         & GSM8K     & 79.2  & 56.2  & 78.9  & \underline{87.9} & \textbf{89.9} \\
                         & SVAMP     & 84.0  &  65.8 & 83.6  & \underline{90.6} & \textbf{92.3}\\
                         & Algebra   & 81.5  & 52.7  & 79.3  & \underline{91.0} &  \textbf{92.8}\\ 
                         & GSM-hard  & 27.5  & \textbf{42.4}  & 27.1  & 35.0 & \underline{36.3} \\
                         & aime24    &  \underline{3.3} & \textbf{6.7}  & \underline{3.3}  & \textbf{6.7}  & \textbf{6.7}\\ \midrule
\multirow{5}{*}{Gemma3 27B} 
                         & GSM8K     & 94.7  & 86.5  & 91.6  & \underline{95.5} & \textbf{96.3} \\
                         & SVAMP     &  94.6 & 92.7  & 94.2  & \textbf{96.5} & \underline{96.2} \\
                         & Algebra   & \underline{98.6}  & 70.7 & 96.4 & \underline{98.6} & \textbf{99.1} \\ 
                         & GSM-hard  & 61.3  &  \underline{65.6} & \textbf{69.2} & 62.0 & 63.1 \\
                         & aime24    & 13.3  & 3.3  & 6.6  & \textbf{23.3} & \underline{20.0}\\ \midrule

\end{tabular}
}
\caption{Solution accuracy across various models and math reasoning datasets. Best performer is shown in bold and second best is underlined}
\label{tab:main_results}
\end{table*}
\section{Experiments}
\subsection{Experimental Setting}
\paragraph{Datasets} 
\begin{table}[H]
\small
\begin{tabular}{lrr}
\toprule
Dataset    & \# Samples \\ \midrule
GSM8K \citep{DBLP:journals/corr/abs-2110-14168}  & 1,319  \\ 
SVAMP \citep{patel-etal-2021-nlp} & 1,000   \\ 
Algebra \citep{DBLP:journals/corr/abs-2304-09102} & 222 \\
GSM-hard \citep{DBLP:journals/corr/abs-2211-10435}  & 1,319 \\
AIME24 \citep{AIME2024} & 30 \\
\bottomrule
\end{tabular}
\caption{Statistics of the datasets we used}
\label{tab:dataset}
\vspace{-5pt}
\end{table}
We conduct our experiments on a diverse collection of five math reasoning datasets, each covering different challenging problem types: GSM8K, SVAMP, Algebra, GSM-hard and AIME24. The GSM-hard dataset is a modified version of GSM8K, where small numerical values have been replaced with larger ones to introduce greater computational difficulty. The details of the statistics of the datasets can be found in Table~\ref{tab:dataset}.

\paragraph{Models} 
For our experiments, we query gpt-4o-mini \footnote{\url{https://openai.com}}, Llama3.1 8b/70b \citep{grattafiori2024llama3herdmodels} and Gemma3 27b \citep{gemmateam2025gemma3technicalreport}.

\subsection{Comparison to state-of-the-art}
In Table~\ref{tab:main_results}, we report solution accuracy across the five math datasets for the four LLMs we tested. We consider three prompting methods as baselines, namely CoT \citep{DBLP:conf/nips/Wei0SBIXCLZ22}, PoT \citep{DBLP:journals/corr/abs-2211-12588} and XoT \citep{liu-etal-2023-plan}, and compare with BloomWise Early Stop (BLES) and BloomWise Majority Voting (BLM). Overall, averaged over all datasets and models, BLM is the top performer, achieving 70.5\% accuracy, followed by BLES at 69.8\%, XoT at 67.5\%, CoT at 66.8\%, and PoT at 60.8\%.
\subsubsection{Performance per task}
BLM is the top performer across almost all datasets, followed closely by BLES, while CoT and XoT achieve comparable performance. More specifically, for GSM8K, BLM achieves 94.2, BLES 93.2, CoT 90.6, XoT 89.4, and PoT 80.2. For SVAMP, the scores are 94.6 for BLM, 94.3 for BLES, 91.2 for CoT, 91.5 for XoT, and 86.2 for PoT. In Algebra, BLM achieves 96.6, BLES 95.5, CoT 92.8, XoT 92.0, and PoT 71.2. In AIME, which is the most demanding in terms of difficulty, BLM and BLES both achieve 15.8, followed by CoT at 12.5, XoT at 9.2, and PoT at 5.0. For GSM-hard, the most challenging dataset in terms of calculation difficulty, methods employing Python programming excel, with PoT being the top performer at 61.5, followed by XoT at 55.2, BLM at 51.1, BLES at 50.1, and CoT at 46.7. These results position our method as the best performer on problems requiring thoughtful reasoning, while methods involving programming excel specifically in calculation-intensive tasks.
\subsubsection{Performance per model}
While BLES and BLM generally achieve the highest performance, the best-performing method varies by model. For GPT-4o-mini, BLM is the top performer with 71.1, followed by BLES at 70.9, XoT at 69.5, CoT at 68.9, and PoT at 67.7. For LLaMA 3.1 70B, XoT achieves the highest score with 74.3, followed by BLM at 72.2, BLES at 70.8, CoT at 70.5, and PoT at 67.1. For LLaMA 3.1 8B, BLM leads with 63.6, followed by BLES at 62.2, CoT at 55.1, XoT at 54.4, and PoT at 44.8. For Gemma3 27B, BLES is the top performer with 75.2, followed by BLM at 74.9, CoT at 72.5, XoT at 71.6, and PoT at 63.8.

\subsection{Trade-offs Between BLM and BLES}
While the top-performing variant varies depending on the model and dataset—meaning there is no universal winner between BLM and BLES—BLM achieves the best overall performance. Nonetheless, BLES presents a compelling alternative when computational efficiency is a priority. Unlike BLM, which requires generating responses for all levels of Bloom’s Taxonomy, BLES terminates the reasoning process as soon as convergence is detected, reducing the number of generated outputs and thus lowering computational cost.
\begin{table*}[t]
\centering
\begin{tabular}{|l|c|c|c|c|c|c|}
\hline
\textbf{Model} & \textbf{Remembering} & \textbf{Understanding} & \textbf{Applying} & \textbf{Analyzing} & \textbf{Evaluating} & \textbf{Creating} \\
\hline
GPT-4o-mini   & 79.3 & 80.1 & 80.3 & 80.7 & 79.2 & 79.6 \\
Llama 3.1 70b & 76.6 & 76.4 & 77.3 & 75.9  & 73.7 & 72.6 \\
Llama 3.1 8b  & 63.2 & 64.2 & 63.4 & 63.3 & 60.4 & 59.1 \\
Gemma 3 27b   & 81.0 & 57.7 & 83.3 & 83.2 & 81.6 & 79.7 \\
\hline
\end{tabular}
\caption{Aggregated scores (\%) per model for each Bloom’s taxonomy level across all datasets ($n = 3{,}890$).}
\label{tab:scores_per_bloom_level}
\end{table*}
\section{Ablation studies and Improved Handling of computations}
The analysis of the results will be structured along two axes. The first concerns the study of the results by Bloom taxonomy level. The second focuses on closing the performance gap between BloomWise and Program-aided Techniques (eg PoT, XoT) on challenging datasets from a computational point of view.
\subsection{Performance at each cognitive level}
For this analysis, we executed the prompts corresponding to each of the levels of the taxonomy (without early stopping). A problem might be correctly solved in more than one levels. 
\subsubsection{LLMs and cognitive abilities}
In Table ~\ref{tab:scores_per_bloom_level}, we show the percentage (\%) of correctly solved
problems per Bloom's level for the LLMs
tested. From this table, we can draw several conclusions about the cognitive skills demonstrated by the LLMs: 

\paragraph{Performance consistency across levels varies significantly among models}: GPT-4o-mini demonstrates homogeneous performance, with scores ranging narrowly between 79.2\% (Evaluating) and 80.7\% (Analyzing), indicating a consistent ability to tackle tasks at any level of cognitive complexity. In contrast, Llama3.1 8b's performance, for instance, ranges from 59.1 to 64.2. Additionally, Gemma 3 27b achieves peak performance in procedural tasks (Applying: 83.3\%; Analyzing: 83.2\%) and factual recall (Remembering: 81.0\%) and performs poorly in Understanding (57.7\%). This might be an indication of a critical weakness in conceptual comprehension despite procedural proficiency, probably implying limited training on such tasks.

\paragraph{Model scale substantially improves overall capability}: The Llama 3.1 70B model outperforms its 8B counterpart by an average margin greater than 10.0\% across all levels, confirming parameter count as a key performance factor.

\paragraph{The best performing level is Applying}: Although the best performing level is not the same across models, Applying is generally the top performer. This behavior is expected due to extensive training on similar methods such as Chain of Thought. 

\paragraph{All models exhibit declining performance at higher taxonomy levels}: all models' performance drops in Evaluating and Creating. This universal trend confirms that these skills remain challenging for the LLMs in mathematical domains, irrespective of scale or architecture. This is probably due to both the inherent difficulty of these cognitive processes and limited training on such tasks.

\subsubsection{Problem type and cognitive depth}
\begin{table*}[ht]
\centering
\begin{tabular}{|c|c|c|c|c|c|c|}
\hline
\textbf{Dataset} & \textbf{Remembering} & \textbf{Understanding} & \textbf{Applying} & \textbf{Analyzing} & \textbf{Evaluating} & \textbf{Creating} \\
\hline
GSM8K     & 90.0 & 82.7 & 90.8 & 91.1 & 88.2 & 88.0 \\
SVAMP     & 91.1 & 88.7 & 90.4 & 90.2 & 90.0 & 87.7 \\
Algebra   & 83.1 & 79.0 & 92.3 & 92.1 & 91.2 & 90.8 \\
GSM-hard  & 48.0 & 41.8 & 49.1 & 48.4 & 45.4 & 44.7 \\
AIME      & 8.4  & 8.4  & 12.5 & 12.5 & 10.0 & 10.0 \\
\hline
\end{tabular}
\caption{Aggregated scores (\%) per dataset for each Bloom’s taxonomy level across all models.}
\label{tab:scores_per_dataset}
\end{table*}

Table~\ref{tab:scores_per_dataset} presents the scores for each dataset and Bloom’s taxonomy level. Several trends and insights emerge from these results.

\paragraph{Performance Across Datasets}: The models achieve the highest performance on the GSM8K and SVAMP datasets, consistently scoring above 87\% across all Bloom’s taxonomy levels. This indicates that current LLMs are highly proficient in standard grade-school math, regardless of the specific cognitive skill being assessed. Performance on the Algebra dataset is also strong, with scores ranging from 79.0\% to 92.3\%, showing that models handle symbolic and procedural tasks well but with slightly more variation than GSM8K or SVAMP. In contrast, GSM-hard and AIME represent a stark drop in accuracy, with all Bloom level scores for GSM-hard in the 41.8\% to 49.1\% range, and for AIME in the 8.4\% to 12.5\% range. These results highlight that state-of-the-art LLMs still struggle substantially with high-complexity, olympiad-style and challenging in terms of calculations problems.
\begin{table*}[!b]
\centering
\begin{tabular}{|c|c|c|c|c|}
\hline
Dataset & BLES & BLM & Program of BLES & Program of BLM\\ \midrule
GSM8K &95.5 & 96.3 & 92.0 & 91.5\\
SVAMP & 96.5 & 96.2 & 94.1 & 93.9\\
Algebra & 98.6 & 99.1 & 89.6 & 88.3\\
GSM-hard & 62.0  & 63.1 & 65.0 & 65.3\\
aime & 23.3 & 20.0 & 20.0 & 23.3 \\
\bottomrule
\end{tabular}
\caption{
Comparison between Program of Bloom and BloomWise-Results for Gemma3 27B}
\label{tab:bloompot}
\end{table*}
\paragraph{Dataset difficulty and patterns}: For the easier datasets from a reasoning perspective—GSM8K, SVAMP, Algebra, and GSM-hard—models achieve their highest accuracy on Applying and Analyzing tasks, indicating strong proficiency with strategies that closely resemble techniques heavily emphasized during training, such as chain-of-thought (CoT) reasoning. Scores for Remembering are also high, suggesting that even pure recall is often sufficient for such problems, but these scores generally do not surpass those for Applying or Analyzing. In contrast, performance consistently dips for Understanding, which emerges as the weakest category in these datasets, and declines moderately for Evaluating and Creating, though these higher-order skills often still outperform Understanding.

For the more difficult dataset, AIME, Remembering and Understanding are the lowest-performing categories, while Applying and Analyzing remain the strongest. Notably, for these challenging problems, Evaluating and Creating sometimes yield better results than lower-order skills (Remembering, Understanding), suggesting that as difficulty and unfamiliarity increase, deeper reasoning may be required for success.

\subsection{Program of Bloom}
In order to reduce errors in calculations, we incorporated Python code into our framework. More specifically, we used the same approaches (BLES and BLM) but modified the prompts to request answers in the form of Python code, which was then safely executed. For the sake of simplicity for this analysis, we only report the results concerning Gemma3 2.7B. 
On the GSM-Hard dataset—the most challenging in terms of calculations—accuracy improved slightly, with gains of 3\% for BLES and 2.2\% for BLM. However, for the remaining datasets, accuracy was considerably lower—or, in the case of AIME, equivalent—compared to the performance achieved with BloomWise (BLES and BLM).
Results can be found in Table \ref{tab:bloompot}. 
A possible explanation for the reduced performance of this variant is that the prompts are not be well-suited to generating answers in code format. The strict structure required for executable code can limit the model's ability to follow the intended prompting, whereas textual rationales provide more flexibility and are often better aligned with the task structure.

\section{Conclusion}
We propose BloomWise, a problem-solving framework that uses prompts inspired by the 
levels of Bloom’s Taxonomy to reach the correct solution. We introduce three variants of our method: EarlyStop (BLES), which halts the process if a solution
is deemed as correct based on a convergence criterion, preventing progression to higher levels of the taxonomy; Majority Voting (BLM), where the final solution is determined by a consensus across the outputs; and Program of Bloom, similar to BLES and BLM but requiring the answer in the form of Python code. Experiments conducted on five math reasoning datasets demonstrated the efficiency of our method, showcasing
not only accuracy-exceeding the state of the art methods- but also providing valuable insights into the cognitive abilities of LLMs. Among the variants, BLM achieved the highest accuracy, while BLES prioritized computational efficiency. The Program of Bloom variant achieved the best accuracy only on the GSM-hard dataset, while it performed the lowest on the rest of the datasets.

Our findings highlight the promise of cognitively-grounded prompting strategies for enhancing LLM performance in zero-shot settings. In future work, we plan to extend BloomWise to additional domains beyond math reasoning, exploring its generalizability and broader applicability.
\section*{Limitations}
We acknowledge that, although our method achieves top performance on most datasets, it struggles with problems that require complex computations. Additionally, while our evaluation focused on mathematics to enable a more targeted analysis, applying our framework to a broader range of tasks would further validate its generalizability and practical utility.

\bibliography{custom}

\appendix

\section{Prompts of BloomWise}
\label{sec:appendix}
In this section, we provide the prompts used for each Bloom's Level (table ~\ref{tab:blooms_prompts}) and the System Prompt (table ~\ref{tab:system_prompt}). 
\begin{table*}[h]
\centering
\begin{tabular}{@{}l p{14cm}@{}}
\toprule
\textbf{Bloom's Level} & \textbf{Prompt} \\
\midrule
 Remembering& You are at the Remembering level. Solve the problem by retrieving, recognizing, and recalling relevant math facts, formulas, definitions, similar problems or the exact same problem from memory. Clearly express what information or problems you recall that is relevant to solving this specific problem. \\
 Understanding& You are at the Understanding level. Solve the problem by constructing meaning from the problem statement and relevant math concepts. Show your thinking by interpreting, exemplifying, classifying, summarizing, inferring, comparing, and explaining the concepts involved and what the problem is asking for.  \\
 Applying&You are at the Applying level. Solve the problem by carrying out or using a known procedure. Clearly show how to apply this procedure to this specific problem step by step.  \\
 Analyzing& You are at the Analyzing level. Solve the problem by breaking it into parts, determining how the parts relate to one another, and identifying patterns or relationships. Show your thought process by differentiating, organizing, and attributing relationships between the math elements. \\
 Evaluating& You are at the Evaluating level. Solve the problem by making judgments about different approaches or potential solutions. Express your thought process by checking, critiquing, and explaining why one approach or answer is better or more appropriate than others. \\
 Creating& You are at the Creating level. Solve the problem by putting together elements to form a new solution strategy or structure. Show your thinking as you generate, plan, or produce a novel approach to this problem.  \\
\bottomrule
\end{tabular}
\caption{Bloom's level Specific Prompts}
\label{tab:blooms_prompts}
\end{table*}

\begin{table*}[!t]
\centering
\begin{tabular}{p{16cm}}  
\toprule
\textbf{System Prompt} \\
\midrule
\vspace{1.5em} \\  
You are a tutor. Solve the given math problem in class using only cognitive skills associated with the specified Bloom's level. Explicitly express your thought process out loud as you solve it, so the student can follow your reasoning.

{Bloom's level specific prompt}

Problem:
{problem}

Provide the final numerical answer at the end in the format: "The final answer is:"
\vspace{1.5em} \\  
\bottomrule
\end{tabular}
\caption{System Prompt}
\label{tab:system_prompt}
\end{table*}

\section{Examples of BloomWise}
\label{sec:appendixB}
In this section, we provide output examples for each level of Bloom’s taxonomy in Tables ~\ref{tab:remembering}, ~\ref{tab:understanding}, ~\ref{tab:applying}, ~\ref{tab:analyzing}, ~\ref{tab:evaluating} and ~\ref{tab:creating}. The problem is stated in Table ~\ref{tab:problem}.  
\begin{table}[h!]
\centering
\begin{tabular}{|p{16cm}|}
\hline
\textbf{Problem} \\
\hline
A candle melts by 2 centimeters every hour that it burns. How many centimeters shorter will a candle be after burning from 1:00 PM to 5:00 PM?
 \\
\hline
\end{tabular}
\caption{Problem Statement}
\label{tab:problem}
\end{table}

\clearpage

\begin{table}[h!]
\centering
\begin{tabular}{|p{4cm}|p{12cm}|}
\hline
\textbf{Level} & \textbf{Solution} \\
\hline
Remembering &  Okay, let's tackle this problem together. We're working at the Remembering level, which means we're going to rely on things we already know: facts, definitions, and maybe similar problems we've seen before. We're not going to do a lot of complex thinking right now. First, let's read the problem carefully: A candle melts by 2 centimeters every hour that it burns. How many centimeters shorter will a candle be after burning from 1:00 PM to 5:00 PM? Okay, what do I remember about solving problems like this? I remember that when we have a rate, something happening per unit of time and a duration of time, we can find the total change by multiplying. That's a key fact! I remember doing problems like \"If a car travels 60 miles per hour for 2 hours, how far does it travel?\". That feels similar. Now, let's identify what we know. The problem tells us the candle melts at a rate of 2 centimeters per hour. That's our rate. Next, we need to figure out how long the candle is burning. It burns from 1:00 PM to 5:00 PM. I remember how to figure out time differences. I can count the hours: 1 PM to 2 PM is 1 hour, 2 PM to 3 PM is another hour, 3 PM to 4 PM is another, and 4 PM to 5 PM is the last hour. That's a total of 4 hours. Alternatively, I remember I can just subtract the start time from the end time: 5 - 1 = 4. So, the candle burns for 4 hours. Now I recall the multiplication fact we discussed earlier. Rate times time equals total change. We have a rate of 2 centimeters per hour, and a time of 4 hours. So, we multiply: 2 centimeters/hour * 4 hours. The hours cancel out, leaving us with centimeters. 2 * 4 = 8. Therefore, the candle will be 8 centimeters shorter. The final answer is: 8 \\
\hline
\end{tabular}
\caption{Example of a response corresponding to Remembering Level}
\label{tab:remembering}
\end{table}

\clearpage

\begin{table}[h!]
\centering
\begin{tabular}{|p{4cm}|p{12cm}|}
\hline
\textbf{Level} & \textbf{Solution} \\
\hline
Understanding & Okay, let's tackle this problem together. We're working at the Understanding level, which means we need to show we get what's happening in the problem, not just jump to a calculation. So, I'm going to talk through my thinking as I go. First, let's interpret the problem. What's the problem telling us? It's telling us about a candle and how it changes over time. Specifically, it says the candle gets shorter. That's important, it's decreasing in length.  The rate of change is key: it melts 2 centimeters every hour. That's a consistent rate, a rule we can use. Now, let's summarize what we need to find out. The problem asks: How many centimeters shorter will a candle be? This means we need to find the total amount the candle decreases in length.  Let's exemplify with a simpler case. If the candle burned for just one hour, how much shorter would it be? The problem tells us 2 centimeters. If it burned for two hours, it would be 2 + 2 = 4 centimeters shorter. See? We're just repeatedly applying that 2 cm/hour rule. Okay, now let's get back to the original problem. We need to figure out how many hours the candle is burning. The candle burns from 1:00 PM to 5:00 PM.  Let's infer the number of hours.  We can count it out: 1 PM to 2 PM is one hour, 2 PM to 3 PM is another hour, 3 PM to 4 PM is another, and 4 PM to 5 PM is the last hour. That's a total of 4 hours. Now we can compare this to our simpler example. We know it melts 2 cm per hour. We now know it burns for 4 hours. So, it's like adding 2 cm four times.  Finally, let's explain how to calculate the total change in length. We'll multiply the rate of melting (2 cm/hour) by the number of hours it burns (4 hours). That's 2 cm/hour * 4 hours. So, 2 multiplied by 4 is 8.  Therefore, the candle will be 8 centimeters shorter. The final answer is: 8  \\
\hline
\end{tabular}
\caption{Example of a response corresponding to Understanding Level}
\label{tab:understanding}
\end{table}

\clearpage

\begin{table}[h!]
\centering
\begin{tabular}{|p{4cm}|p{12cm}|}
\hline
\textbf{Level} & \textbf{Solution} \\
\hline
Applying &  Alright, let's tackle this problem together. We're working at the 'Applying' level, which means we need to use a procedure we already know to solve this. In this case, the procedure is understanding rate and using multiplication. First, let's identify what we know. We know the candle melts at a rate of 2 centimeters per hour. That's our key piece of information. Next, we need to figure out how long the candle is burning. The problem tells us it burns from 1:00 PM to 5:00 PM.  Let's count the hours: 1:00 PM to 2:00 PM is one hour, 2:00 PM to 3:00 PM is another hour, 3:00 PM to 4:00 PM is another hour, and 4:00 PM to 5:00 PM is the final hour. So, that's a total of 4 hours. Now we can apply the procedure. We know the rate (2 cm/hour) and the time (4 hours). To find the total change in length, we multiply the rate by the time. So, we'll do: 2 centimeters/hour * 4 hours. The 'hour' units cancel out, leaving us with centimeters. 2 * 4 = 8. Therefore, the candle will be 8 centimeters shorter. The final answer is: 8 \\
\hline
\end{tabular}
\caption{Example of a response corresponding to Applying Level}
\label{tab:applying}
\end{table}

\clearpage

\begin{table}[h!]
\centering
\begin{tabular}{|p{4cm}|p{12cm}|}
\hline
\textbf{Level} & \textbf{Solution} \\
\hline
Analyzing & Alright class, let's tackle this candle problem. We're working on analyzing today, which means we need to break this problem down into its pieces and see how they connect. Don't just jump to a calculation,  let's understand what's happening first. Okay, what information are we given? We know the candle melts at a rate 2 centimeters per hour. That's important. That's our key relationship.  We also know a time period: from 1:00 PM to 5:00 PM. Now, let's differentiate. We have a change in length (how much shorter the candle gets) and a change in time (the duration it burns).  Our rate connects these two.  Melting rate is essentially change in length divided by change in time. Let's organize this.  We need to figure out the total change in time.  How many hours are there between 1:00 PM and 5:00 PM?  Let's see: 1:00 to 2:00 is one hour, 2:00 to 3:00 is another, 3:00 to 4:00, and 4:00 to 5:00. That's four hours.  So our change in time is 4 hours. Now, let's attribute the relationship. We know the candle melts 2 centimeters for every hour.  That 'for every' signals a multiplicative relationship.  We can set up a sort of proportion, even though we don't need to formally write it out as one.  If it melts 2 cm in 1 hour, then in 4 hours, it will melt, well, four times as much, right? So, we are multiplying the rate of melting (2 cm/hour) by the number of hours (4 hours). This will give us the total change in length. 2 cm/hour * 4 hours = 8 cm. Therefore, the candle will be 8 centimeters shorter. The final answer is: 8 \\
\hline
\end{tabular}
\caption{Example of a response corresponding to Analyzing Level}
\label{tab:analyzing}
\end{table}

\clearpage

\begin{table}[h!]
\centering
\begin{tabular}{|p{4cm}|p{12cm}|}
\hline
\textbf{Level} & \textbf{Solution} \\
\hline
Evaluating & Okay, let's tackle this candle problem. We're at the 'Evaluating' level today, which means we're not just doing the math, but thinking about how we're doing it and whether our approach makes sense. First, let's quickly identify what the problem is asking. It wants to know the total amount the candle shrinks after burning for a certain period. Seems straightforward. Now, a really quick, almost automatic thought might be to just multiply 2 centimeters (the melt rate) by 5 (because 5 PM is 5 hours after 1 PM). That would give us an answer, and it's a common way to approach rate problems. Let's hold that thought 2 x 5 = 10. But, let's evaluate that. Does that just automatically feel right? Yes, it seems logical, but we should always double-check our assumptions. The problem clearly states the candle melts every hour it burns. We need to make sure we've correctly calculated the number of hours. Looking at the times, 1:00 PM to 5:00 PM. Let's think through it. 1 PM to 2 PM is one hour. 2 PM to 3 PM is another. 3 PM to 4 PM, and 4 PM to 5 PM. That's four hours, not five.  So, that initial quick calculation of multiplying by 5 was incorrect. That's a good example of why evaluating is important! We caught a potential mistake. Okay, so let's revisit. We have a melt rate of 2 cm/hour and a burn time of 4 hours. Now, multiplying 2 cm/hour by 4 hours seems much more solid. Let's think about units too, which is a good evaluating habit.  Centimeters per hour * hours- the hours cancel out, leaving us with just centimeters. That makes sense- we want an answer in length! So, 2 * 4 = 8. Let's quickly critique again. Does 8 centimeters seem reasonable? If a candle melts 2cm an hour, and burns for four hours, 8cm seems perfectly plausible. It's a manageable number, not ridiculously large or small. Therefore, after burning from 1:00 PM to 5:00 PM, the candle will be 8 centimeters shorter. The final answer is: 8  \\
\hline
\end{tabular}
\caption{Example of a response corresponding to Evaluating Level}
\label{tab:evaluating}
\end{table}

\clearpage

\begin{table}[h!]
\centering
\begin{tabular}{|p{4cm}|p{12cm}|}
\hline
\textbf{Level} & \textbf{Solution} \\
\hline
Creating & Okay, let's tackle this candle problem! Now, we're aiming for the Creating level of Bloom's Taxonomy, which means we aren't just recalling facts or applying a formula directly. We need to build a solution. I want to think beyond just \"hours times melt rate.\" While that is part of it, let's see if we can frame this in a slightly more robust way, maybe thinking about how we represent time and change. First, let's identify what we know. We know the candle melts at a constant rate: 2 centimeters per hour. That's our core piece of information.  We also know the start and end times: 1:00 PM and 5:00 PM.  Now, usually, we 'd jump to subtraction, right? 5 minus 1 is 4. But let's pause and think... what is subtraction actually doing in this context? It's finding the difference between two points in time.  That difference represents the duration of burning. So, instead of immediately calculating 5-1, let's explicitly define a variable. Let's say 't' represents the burning time in hours.  We can create a little formula for 't': t = End Time - Start Time. Okay, that's a bit abstract for this problem, honestly, but it's getting us to think about the components.  Now we can plug in our numbers: t = 5:00 PM - 1:00 PM = 4 hours. See? We didn't just know to subtract; we defined the subtraction as a way to calculate the burning time. That's a small step towards creating a more generalized approach. Now, we know the burning time, and we know the melt rate.  Let's create another relationship. Let 'L' represent the total length the candle melts. We can say: L = Burning Time * Melt Rate, L = t * 2 cm/hour. Again, this seems obvious, but we're building the logic from first principles, which is what 'Creating' is about. Now, we can substitute the value of 't' we found earlier:L = 4 hours * 2 cm/hour = 8 cm. So, the candle will be 8 centimeters shorter. We didn't just apply a formula; we constructed the formulas to represent the relationships in the problem.  That's the difference between applying and creating. We could have just done 5-1=4, then 4*2=8, but this way we 've explicitly broken down why we 're doing each step. The final answer is: 8  \\
\hline
\end{tabular}
\caption{Example of a response corresponding to Creating Level}
\label{tab:creating}
\end{table}

\end{document}